\documentclass[10pt]{article} % For LaTeX2e
\usepackage[preprint]{tmlr}
\usepackage{hyperref}
\usepackage{url}
\usepackage{xspace}
\usepackage{listings}
\usepackage{xcolor}
\definecolor{backcolour}{rgb}{0.95,0.95,0.92}
\usepackage{graphicx}
\usepackage{amsmath}
\usepackage{multirow}

\newcommand{\ie}{\mbox{i.e.,}\xspace}
\newcommand{\eg}{\mbox{e.g.,}\xspace}
\newcommand{\etc}{\mbox{etc.,}\xspace}
\newcommand{\xhdr}[1]{{\noindent\bfseries #1}.}

\newcommand{\tool}{FoCusNet\xspace}
\newcommand{\toolLong}{FoCusNet (Focused Constraints Net)\xspace}

\newcommand{\task}{Words Checker\xspace}
\newcommand{\taskAcr}{Words Checker\xspace}
\newcommand{\problem}{Large-Scale Constraint Generation\xspace}
\newcommand{\problemLong}{Large-Scale Constraint Generation (LSCG)\xspace}
\newcommand{\problemAcr}{LSCG\xspace}
\newcommand{\tssFull}{Test Steering Strategies (TSS)\xspace}
\newcommand{\tss}{Test Steering Strategies\xspace}
\newcommand{\tssAcr}{TSS\xspace}

\newcommand{\LLMs}{Large Language Models (LLMs)\xspace}
\newcommand{\CoT}{Chain of Thought\xspace}
\newcommand{\CoTFull}{Chain of Thought (CoT)\xspace}
\newcommand{\CoTAcr}{CoT\xspace}

\newcommand{\RAG}{Retrieval Augmented Generation (RAG)\xspace}

\lstnewenvironment{prompt}{%
    \thicklines
    \lstset{
        backgroundcolor=\color{backcolour},
        rulecolor=\color{backcolour!40!black},
        frameround=tttt,
        frame=single,
        basicstyle=\ttfamily\normalsize,
        basewidth=0.50em,
        keywordstyle=\bfseries,
        xleftmargin=.02\linewidth,
        xrightmargin=.02\linewidth,
        escapeinside={<@}{@>},
        numbers=none
}}{}

\title{\textit{\problem}\\Can LLMs Parse Hundreds of Constraints?}

\author{\name Matteo Boffa \email matteo.boffa@polito.it \\
      Politecnico di Torino
      \AND
      \name Jiaxuan You \email jiaxuan@illinois.edu \\
      University of Urbana Champaign (UIUC)
      }

\begin{document}
\maketitle
\begin{abstract}
%%%%%%%%%%%%%%%%%%%%%%%%%%%%%%%%%%%%%%%%%%%%%%%%%%%% Starting point, but too long 

%The AI revolution, led by LLMs, is transforming industries and surpassing human performance in tasks like text understanding and image classification. As research advances toward more complex challenges (e.g., mathematics, planning), this work takes a step back to explore how models handle an increasing number of simple constraints.
%In particular, we benchmark the performance of state-of-the-art LLMs on a simple string-check problem: Given a list of forbidden words and a sentence, does the sentence contain any of the forbidden words? We systematically stress increasingly larger models -- from 8B to 70B -- with an increasing number of words -- from 100 to 1000 -- and show that, regardless of the model size, all models experience an $\approx20\%$ accuracy drop when the number of words becomes too large.

%This result shows that, despite showing impressive results in very complicated objectives, LLMs fall short on trivial tasks when the number of constraints becomes too high. We believe that, unlike humans, current architectures are still not capable of successfully approximate a large number of constraints into fewer high-level concepts. Inspired by this intuition, we also explore a solution that enforces hierarchical reasoning by clustering the embedding space to group similar constraints together. Although not yet successful, we show that this proposal yields promising results, possibly paving the way for future breakthroughs. 

Recent research has explored the \textit{constrained generation} capabilities of \LLMs when explicitly prompted by few task-specific requirements. In contrast, we introduce \textit{\problemLong}, a new problem that evaluates whether LLMs can parse a large, fine-grained, generic list of constraints. To examine the LLMs' ability to handle an increasing number constraints, we create a practical instance of \problemAcr, called \textit{\task}. In \task, we evaluate the impact of model characteristics (\eg size, family) and steering techniques (\eg \textit{Simple Prompt}, \textit{Chain of Thought}, \textit{Best of N}) on performance. We also propose \textit{\tool}, a small and dedicated model that parses the original list of constraints into a smaller subset, helping the LLM focus on relevant constraints. Experiments reveal that existing solutions suffer a significant performance drop as the number of constraints increases, with \tool showing an 8-13\% accuracy boost.
% the most robust and promising results.  

%Our findings highlight the importance of \problemAcr and call on the research community to further explore this challenge by developing new benchmarks. To this end, we release our code and datasets.
\end{abstract}

%%%%%%%%%%%%%%%%%%%%%%%%%%%%%%%%%%%%%
\section{Introduction}\label{sec:intro}
%Recent research has extensively explored the ability of \LLMs to complete tasks following instructions, that is, explicit directives on how the task should be performed.
Instructions are prompts or directives, written in natural language, that guide the model to perform a specific task~\cite{ouyang2022training}. The recent literature has extensively studied the ability of \LLMs to follow instructions requiring complex reasoning~\cite{wang2023how}, focusing on multiple requirements and for multiple rounds~\cite{10.1609/aaai.v38i16.29777, he2024from}, and even dealing with long texts~\cite{bai-etal-2024-longbench, li2024needlebenchllmsretrievalreasoning}.  
To address the real-world urgency for controllable outputs~\cite{10.1145/3613905.3650756, 10.1145/3663529.3663849}, researchers have also investigated whether LLMs, provided with clear indications of the expected answer, could support \textit{constrained generation} (\eg ``the answer must contain exactly N words'')~\cite{sun-etal-2023-evaluating, yao2024collie, xia-etal-2024-fofo}.
% pmlr-v70-hu17e,

\begin{figure*}[tb]
    \centering
    \includegraphics[width=\linewidth]{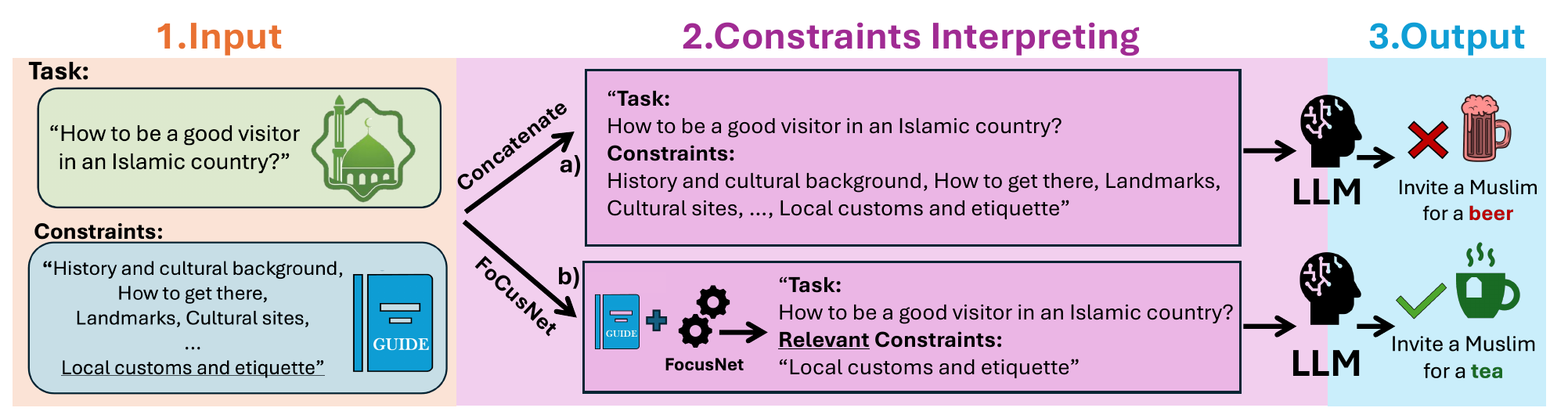}
    \caption{In \problemAcr, the model must generate a \textbf{valid answer} while adhering to an \textbf{input task} and a \textbf{long list of constraints}. In the example, this can be done either by (a) directly interpreting the \textbf{concatenated} task and constraints or (b) using a \textbf{\tool} to \textbf{extract relevant constraints}. The first approach may lead to \textbf{inappropriate responses} (\eg offering beer to a Muslim \cite{naous2024having}), while the second ensures \textbf{valid answers}.}
    \label{fig:figure_1}
    \vspace{-4mm}
\end{figure*}

In this paper, we take a step further in defining instruction-following tasks. Instead of the few task-specific indications the literature has used so far, we focus on scenarios with a high number of fine-grained but general constraints that the model must respect to generate a valid answer. Consider the example in Fig.~\ref{fig:figure_1}. The model faces a social task (\eg ``be a good visitor in an Islamic country''), and can access a comprehensive travel guide with generic information on how to achieve the goal (\ie long list of constraints). Could the LLM, with the sole aid of the generic travel guide and no other explicit instruction, realise that ``inviting a Muslim for a beer after prayer''~\cite{naous2024having} is not a good way to solve the task?

We call this new framework \textit{\problemLong}. \problemAcr examines whether LLMs can replicate humans' \textit{practical intelligence}~\cite{Sternberg1986-STEBIA-5}, \ie the ability to interpret and adapt to the context. In particular, facing \problemAcr the model is not tasked to solve complex reasoning problems, but rather i) to consult broad and generic guidelines (\eg travel guide, but also updated documentation while coding~\cite{10.1145/3656296, 299699}), ii) to identify the requirements relevant for the specific problem, and iii) to apply them to derive a valid solution. 

As it is currently unclear whether and how LLMs' capabilities could scale with the hundreds (if not thousands) of constraints that a travel guide or some code documentation could provide, we implement a concrete instance of \problemAcr, \textit{\task}. 
We design \taskAcr as a simple problem, not requiring particular reasoning skills, to explicitly study how the performance of LLMs while solving the task is affected by the number of constraints. In \taskAcr, the model is given as input a list of forbidden words and a sample sentence. The task is to classify the sentence as \textit{valid} (\ie does not contain forbidden words) or \textit{invalid} (\ie contains at least one forbidden word). 

We create different instances of \taskAcr with increasingly larger lists of forbidden words (\eg 100, 500 and 100). Then, we systematically evaluate how features such as model family -- Meta's \textit{LLama}~\cite{grattafiori2024llama3herdmodels} vs. Deepseek's \textit{R1}~\cite{deepseekai2025deepseekr1incentivizingreasoningcapability}),  size -- 8B vs. 70B, and \tssFull -- \textit{Simple Prompt}, \textit{\CoT}~\cite{10.5555/3600270.3602070, lightman2024lets} and \textit{Best of N}~\cite{chen2024teaching, madaan2023selfrefine} affect the results. 

Furthermore, inspired by \textit{\RAG}~\cite{lewis2020retrieval-augmented} and the recent literature~\cite{DBLP:journals/corr/abs-2110-14168, shi2024replug}, we propose \textit{\toolLong}, a lightweight and customizable model to parse the originally large list of constraints into a smaller set of constraints relevant to the task, helping the LLM to better focus.
In \taskAcr, \tool is a $\sim300$k parameters model that we train to determine whether a set of words is present in a sentence. During inference, it preprocesses the long list of forbidden words and parses it into a smaller set of potential suspects, allowing the LLM to focus more effectively on meaningful instances.

\begin{figure}[t]
    \centering
    \includegraphics[width=0.6\linewidth]{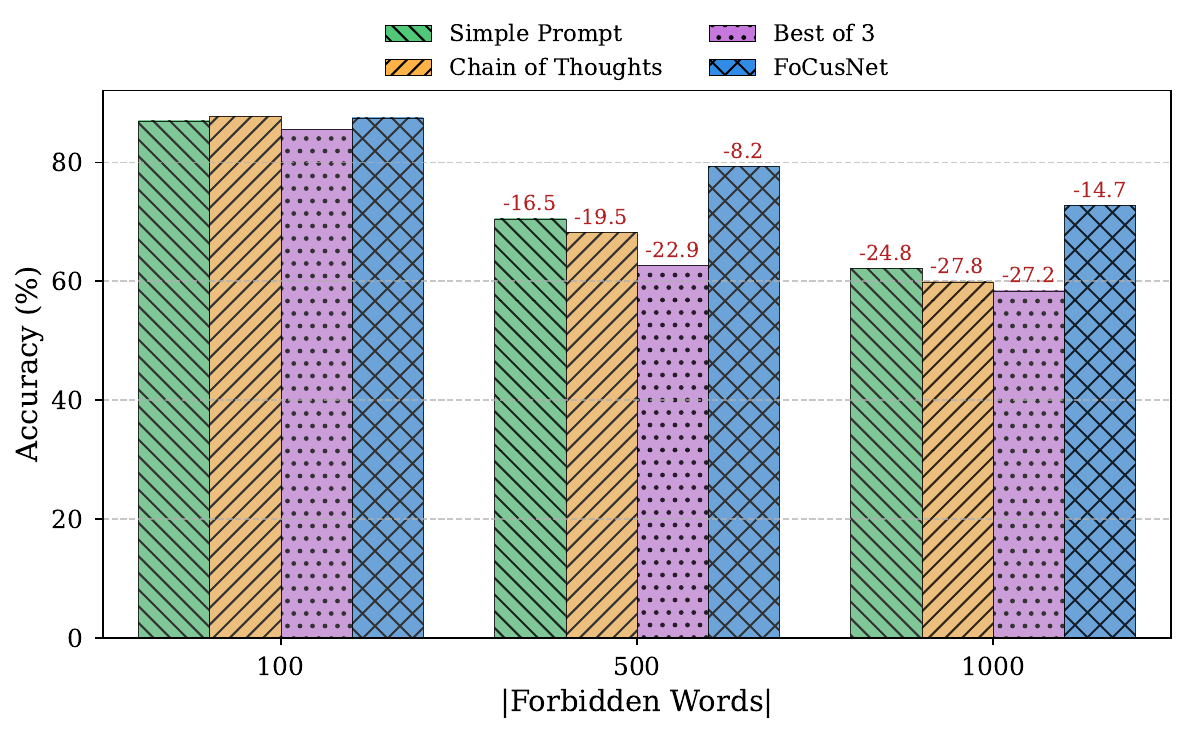}
    \vspace{-3mm}
    \caption{\tool significantly outperforms typical LLM inference methods on the proposed \task task (\textit{DeepSeek-R1-Distill-Llama-8B}). Red numbers indicate differences compared to 100-word scenario.
    }
    \vspace{-3mm}
    \label{fig:figure_2}
\end{figure}

The results of a distilled 8B LLM in \taskAcr, shown in Fig.~\ref{fig:figure_2}, are striking: traditional \tss, including simple prompting, suffer a drastic performance drop -- down to $\sim27.8\%$  accuracy. Manual analysis reveals that the model often processes words individually, losing focus, and sometimes conflating its reasoning process with the actual task. For example, it may incorrectly assert that a word is present simply because it appears in a self-generated list. 

Our approach proves the most robust, leveraging the synergy between two models. \tool, trained to detect the presence of words with accuracy 90\%, effectively narrows the search space (\ie average of 30 suspicious words out of 1000). The LLM, in turn, benefits from this reduced scope, filtering out false positives from \tool and improving overall accuracy. In sum, our contributions are:

\textendash{}~\textbf{\problem}: A novel problem to evaluate the ability of current LLMs to automatically parse a large number of constraints and identify the relevant ones.

\textendash{}~\textbf{\task}: A practical example of \problemAcr where the model identifies invalid sentences as the number of forbidden words increases. We systematically experiment 2 models (\textit{LLama} and \textit{R1}), 2 model sizes (8B and 70B), and 3 \tssAcr (prompt-based, CoT, Best of N). 

\textendash{}~\textbf{\tool}: A small dedicated model that works in conjunction with the LLM, helping it to better focus on relevant constraints.
% generic model we add to the experimental plane. \tool employs a

\textendash{}~\textbf{Code and Datasets}: To reproduce \task and \tool and help the community benchmarking \problemAcr\footnote{\url{https://github.com/ulab-uiuc/Large_Scale_Constraint_Generation}}.

\section{Related Work}\label{sec:lit_review}
% \subsection{Instruction-Following}

% - Rule following
% - Constrained Generation

\xhdr{Instruction-Following abilities of LLMs} The challenge of constraining textual generation has been studied since the early days of NLP~\cite{pmlr-v70-hu17e}, but the rise of LLMs has dramatically increased expectations beyond merely ``producing plausible text''~\cite{brown2020language, wei2022emergent}. Modern LLMs are expected to follow complex instructions, handle multiple constraints across interactions~\cite{10.1609/aaai.v38i16.29777, he2024from}, and process long texts~\cite{bai-etal-2024-longbench, li2024needlebenchllmsretrievalreasoning}.
Yet, this problem remains unsolved. Studies show that LLMs struggle with adherence to rules~\cite{mu2024can}, format following varies widely across domains~\cite{xia-etal-2024-fofo}, open-source models are still behind closed source solutions~\cite{wang2023how} and smaller models still perform poorly in structured tasks~\cite{wang2025verifiable}.
Most of the previous evaluations assume interactive chat-like settings, with few clear user instructions specific to the required task. In contrast, we contribute to this line of research by examining how LLMs perform when given an extensive list of fine-grained yet generic requirements to satisfy.

%While human evaluation can be accurate, it suffers from subjective bias, high costs, and a lack of reproducibility~\cite{ouyang2022training}. 
%LLM-based evaluation methods offer more scalable and robust alternatives for assessing instruction following performance~\cite{he2024complex, qin2024infobench, zeng2024evaluating}. 

\xhdr{Instruction Tuning} Given these challenges, instruction tuning might seem like a natural candidate for improving adherence to complex and fine-grained constraints. Prior work has highlighted its role in enhancing generalization capabilities~\cite{chung2022scaling, mishra2022cross-task, thoppilan2022lamda}, and even a small set of high-quality instructions can lead to performance gains~\cite{zhou2023lima, chen2024alpagasus}. However, despite well-established guidelines for crafting such instructions~\cite{10.5555/3692070.3694581, he2024complexsimpleenhancingmulticonstraint, zhang-etal-2024-recost}, instruction tuning remains costly and resource-intensive. This makes it unsuitable for large-scale applications that require customization~\cite{10.1145/2959100.2959153, 10.1561/1500000066}, continuous knowledge updates~\cite{lewis2020retrieval-augmented}, or, like our example in Fig.~\ref{fig:figure_2}, cultural adaptation~\cite{adilazuarda-etal-2024-towards, 10.1145/3582269.3615599}. Instead, we argue that LLMs should, like humans, handle unfamiliar constraints by leveraging external knowledge sources while relying on their reasoning abilities to interpret and respond accordingly. Consequently, we do not employ instruction tuning to further specialize our models.

\xhdr{Test Steering Strategies} Rather than modifying a model through instruction tuning, an alternative approach is to guide LLM outputs at inference using test-time steering strategies. These methods enhance rule adherence without the cost and inflexibility of fine-tuning. Prior research has explored various controlled generation techniques to enforce constraints~\cite{hu2018toward}.
LLMs have shown strong performance with simple interventions like Chain-of-Thought (CoT) prompting~\cite{wei2023chain-of-thought}. However, studies suggest that such methods alone may be insufficient for handling fine-grained, hard constraints~\cite{sun-etal-2023-evaluating}.
To address this, researchers have investigated best-of-$K$ selection~\cite{nakano2022webgpt, stiennon2020learning}, where multiple independent samples are generated, scored, and ranked to select the most suitable output. Other approaches include rejection-sampling-based methods~\cite{liu2024statistical}, reward-model-guided decoding~\cite{yang2021fudge, deng2023reward-augmented}, and constraint-aware streaming algorithms~\cite{krause2021gedi, liu2021dexperts}.
Building on this body of work, we assess the rule-following capabilities of LLMs using various test-time steering strategies.

\xhdr{Auxiliary Modules for LLMs} In this paper, we present \tool, a modular support model that enhances LLMs' ability to follow constraints. Unlike base model modifications, \tool acts as an auxiliary module that identifies and prioritizes relevant constraints, guiding the LLM’s generation process. It provides an intermediate solution between resource-heavy instruction tuning and simpler test-time steering methods, which, while more efficient, may struggle with complex tasks.

Similar approaches using specialized support models for LLMs have been explored in various text generation tasks. For example, retrieval-augmented generation (RAG) \cite{lewis2020retrieval-augmented, shi2024replug} improves LLM responses by incorporating external knowledge, while classifier-based safeguards promote responsible generation \cite{sharma2025constitutional}. Furthermore, researchers have also developed classifier-based content moderation systems \cite{chi2024llama, inan2023llama, rebedea2023nemo} and output filtering techniques to address jailbreak vulnerabilities \cite{kim2024testing},

% Additionally, \tool's architecture can be tailored to the nature of the constraints: for example, a \textit{GNN} can be employed for graph-structured constraints, while a \textit{Transformer} may be more suitable for textual constraints.
%Eventually, even with simpler architectures, such as an attention-based network, the model can learn to group constraints into higher-level concepts within its embedding space, mirroring human cognitive processes where individual episodes are abstracted into broader concepts~\cite{Snell2013-SNETDO-4, 10.5555/2534455}. 

% We will provide an example in Section~\ref{sec:results}.

%%%%%%%%%%%%%%%%%%%%%%%%%%%%%%%%%%%%%
\section{\problem}\label{sec:problem_def}
In this Section, we formally define \problemAcr, relate \tss techniques with \problemAcr and finally introduce \tool.  
\subsection{Formal Definition}
In constrained generation, LLMs autoregressively generate an output sequence $y$ according to an input task $t$ and a set of constraints $c=\{c_1, c_2, \dots, c_C\}$. \problemAcr is a specific case of constrained generation characterized by a large number of constraints (\ie $C \geq 100$). We suppose both $t$, and the constraints $c_i$ with $i\in C$ to be string-based. Although this assumption does not cover the most general case (see Sect.~\ref{sec:disc_limit}), it is sufficient to model real-world scenarios such as the travel guide and documentation examples of Sect.~\ref{sec:intro}. 

We define the LLM input \textit{query}: $q=e(t) \parallel p(c)$, where $\parallel$ is the concatenation. Specifically, here $e$ and $p$ are \textit{\tss} that can be applied to improve model performance: $e$ is a function that \textit{enhance} the definition of the task, while $p$ helps \textit{parsing} the constraints. We provide more details in the next section. 

We represent the LLM as a function $f_\theta:q\to y$. This means that the LLM generates an answer $y$ as $y=f_\theta(q)$ according to its pre-trained weights $\theta$. A model-generated answer $y$ is valid for a given query $q$ if it correctly solves the task $t$ while adhering to the constraints $c$.

\subsection{Existing \tss}
Here, we list the most prominent \tssAcr previously identified in the literature and examine how they apply in our formulation. We provide a summary in Tab~\ref{tab:schema_test_steering}.

\begin{table}[t]
\centering
%\vspace{-5mm}
\caption{Summary of how different steering solutions produces the final query $q=e(t) \parallel p(c)$. }
%\resizebox{\columnwidth}{!}{%
\vspace{2mm}
\begin{tabular}{ccc}
\hline
\textbf{Test steering} & \textbf{Enhance - $e(t)$} & \textbf{Parse - $p(c)$}                 \\ \hline
Simple Prompt          & $t$                       & $c_1 \parallel c_2 \parallel \dots \parallel c_C$ \\
Chain of Though        & $t \parallel g$              & $c_1 \parallel c_2 \parallel \dots \parallel c_C$ \\
Best of N              & $t \parallel g$              & $y_1 \parallel y_2 \parallel \dots \parallel y_N$ \\
\tool                  & $t \parallel g$              & $f_\phi(c)$                             \\ \hline
\end{tabular}%
%}
%\vspace{-3mm}
\label{tab:schema_test_steering}
\end{table}

\xhdr{Simple Prompt} As both $t$ and $c$ are text-based, a natural approach is to simply \textit{concatenate} them: $q = t \parallel c_1 \parallel c_2 \parallel \dots \parallel c_C$. 

\xhdr{\CoTFull} To enhance the reasoning capabilities of the LLM, we modify $t$ by appending a guide phrase $g$, such as ``\textit{Think step by step}'': $q = t \parallel g \parallel c_1 \parallel c_2 \parallel \dots \parallel c_C$. 

%\xhdr{Divide and conquer} Since \problemAcr focuses on problems with a large number of constraints $c$, a possible strategy is to first reduce them to a more relevant subset before generating the answer. In this case, the LLM first selects a subset $k$ by applying $k=f_\theta(c)$; then follows the \CoTAcr strategy to obtain $q = t \parallel g \parallel k$. In essence, the LLM acts as the parsing function, \ie $f_\theta(c) \sim p(c)$.

\xhdr{Best of N} Finally, to improve the interpretation of the $C$ constraints, we can involve a panel of $N$ judges (\eg independent runs of the model), each performing \CoTAcr reasoning independently, followed by a recap step to produce the final answer. Formally, let $y_{n}=f_{n,\theta}(t\parallel g\parallel c_1 \parallel c_2 \parallel \dots \parallel c_C)$ denote the answer of the nth judge, where $n\in N$. Then, we can aggregate all the responses into a refined query: $q = t \parallel y_1\parallel y_2 \parallel \dots \parallel y_N$.

\subsection{\tool}

\xhdr{Definition} Here, the goal is to learn an approximation of $p(c):c \rightarrow k$ to reduce the large set of $C$ constraints $c$ to a more compact subset $k\in K$ of relevant constraints. To do that, we introduce a dedicated model, \tool. Specifically, we define \tool as a function $f_\phi$ with learnable parameters $\phi$, trained on task-specific data to filter relevant constraints. Once trained, \tool applies this filtering as $k = f_\phi(c)$, which yields the final query formulation: $q = t \parallel g \parallel k \parallel$.

\xhdr{Training \tool} We train \tool to perform a binary classification task over individual constraints. Specifically, \tool operates on triplets $(\hat{c}, s, l)$. Here, $\hat{c}=\{c_1, c_2, \dots, c_M\}$ is a subset of $M$ constraints from $c$; $s$ is a text-based instance where the constraint is satisfied or violated, and $l \in \{0,1\}$ is a label indicating whether the constraint is violated (1) or not (0).
For example, consider Fig.~\ref{fig:figure_1}. The set of constraints is \{$c_1=$``\textit{Respect local customs and etiquette when visiting an Islamic country}''\}; the instance is $s=$ ``\textit{Invite a Muslim for a beer}''; the corresponding label $l$ is violated ($l=1$). 

\xhdr{Inference with \tool} During inference, \tool receives as input the tuple of constraints and task ($c$, $t$) and generates a \textit{relevance mask}, $m = \{m_1, m_2, \dots, m_C\}$ with $m_i \in \{0, 1\}$ and $i \in C$. The mask determines which constraints are relevant for the task. Applying the mask yields the reduced set: $k=\{c_i \mid m_i =1,\forall i\ \in C \}$. 

As in any alerting system, \tool aims at compromising \textit{recall} and \textit{precision}. Ideally, we would like \tool to reduce the number of false positives, \ie irrelevant constraints mistakenly included. In fact, a large number of false positives leads to a larger and noisy set $k$. At the same time, it is essential to minimize false negatives, as excluding relevant constraints could hinder the LLM’s ability to generate valid outputs.

%%%%%%%%%%%%%%%%%%%%%%%%%%%%%%%%%%%%%
\section{Methodology}\label{sec:methodology}
In this section, we discuss the engineering of \task and, consequently, \tool's training. 

\begin{figure*}[!t]
    \centering
    \vspace{-10mm}
    \includegraphics[width=\linewidth]{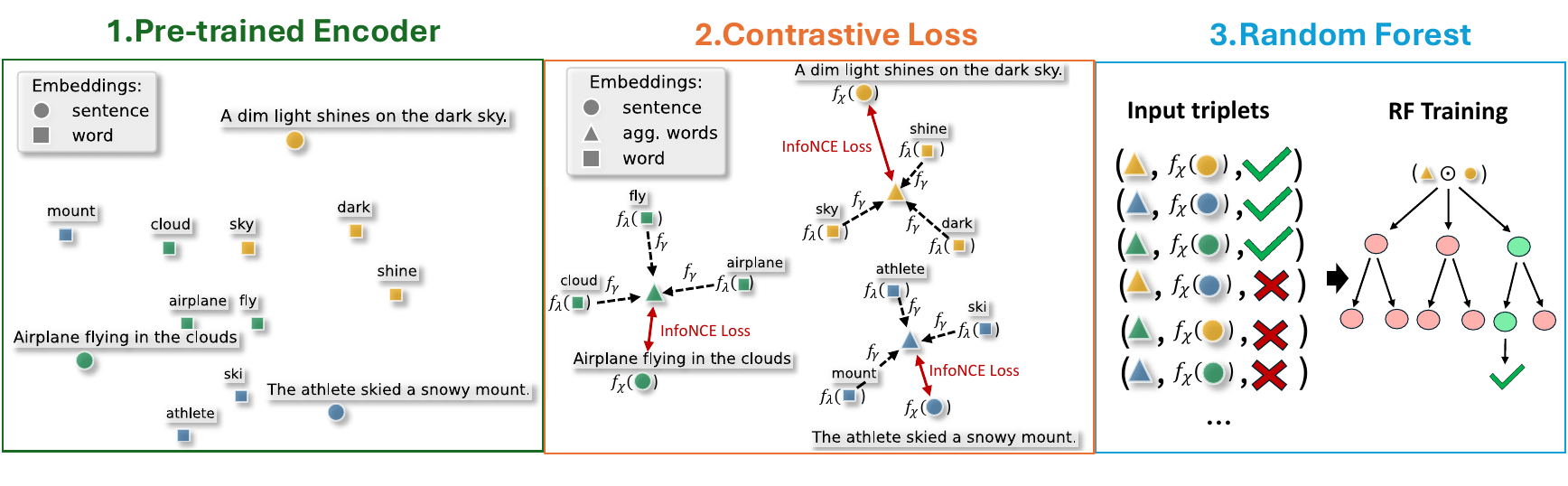}
    \vspace{-8mm}
    \caption{Training pipeline of \tool for \task. The model receives as input a batch of sentences and words. In Phase 1, \tool uses a \textbf{frozen pre-trained model} to map the input into sentences (circles) and words (squares) embeddings. Then, in Phase 2, \tool learns to \textbf{refine the sentence embeddings} ($f_\chi$) and to \textbf{aggregate the words embeddings} ($f_\gamma$, $f_\lambda$) with a InfoNCE contrastive loss. Eventually, in Phase 3 \tool train a Random Forest to \textbf{discriminate positive and negative examples}.}
    \vspace{-1mm}
    \label{fig:model_schema}
\end{figure*}
\subsection{\task}
\xhdr{Problem Definition} \task is an instance of \problemAcr, where an LLM must classify a sentence as \textit{valid} or \textit{invalid} based on a dynamically provided list of forbidden words. Formally, given a sentence $S = (w_1, w_2, \dots, w_n)$ and a set of forbidden words $F = \{ w_{f1}, w_{f2}, \dots, w_{fm} \}$, the model must determine whether $S$ contains any word morphologically related to an element of $F$. A sentence is classified as \textit{invalid} if $\exists w_{fi} \in F$ such that $w_{fi}$ is a root or morphological variant of any $w_j \in S$, and \textit{valid} otherwise. For example, given the sentence ``The athlete skied a snowy mountain'' and $F=\{\text{ski}\}$, the output should be \textit{invalid}, since ``skied'' is a morphological variant of ``ski''. In contrast, for ``The bathroom has recently been cleaned'' and $F=\{\text{restroom}\}$, the output should be \textit{valid}, as no word in $S$ morphologically relates to ``restroom''.

\xhdr{Ratio behind \task} We explicitly design \task to study the impact of an increasing number of forbidden words on LLM performance. Therefore, unlike other constrained generation problems, this task does not require complex reasoning. Instead, we engineer \task as a simple problem that an advanced, morphologically aware string-matching algorithm -- without concern for synonyms -- could potentially solve. In summary, \task serves as an in vitro study on \problemAcr.
At the same time, \task has practical applications. Consider a scenario where $S$ is an LLM-generated response $y$ in a conversation, and $F$ consists of words the user explicitly wants to avoid (\eg when paraphrasing text, for secret keeping, \etc).

\xhdr{Testing Dataset}
To construct a dataset for \task, we use the \textit{CommonGen}~\cite{lin2020commongen} benchmark, originally designed for traditional constrained text generation. Each entry in CommonGen consists of a sentence and a variable-sized list of $W$ words that are morphologically present in it. For example, an entry may contain ``The athlete skied a snowy mountain'' with the corresponding words [``ski'', ``snow'']. 

We derive our dataset from two partitions of CommonGen, namely the \textit{challenge train sample} and \textit{challenge validation sample}\footnote{The test partitions of CommonGen do not contain reference sentences.}. For these partitions, $W$ ranges from 1 to 4. 
Given a pool size of candidate forbidden words $|F|$, we:
i) construct a vocabulary from all CommonGen partitions, and
ii) iterate over the selected partitions to generate valid and invalid samples.
To create an \textit{invalid} example, we retain $W$ CommonGen words and randomly sample $|F| - W$ additional vocabulary words. For a \textit{valid} example, we select $|F|$ random words ensuring that none is morphologically present in the sentence.

We generate four versions of \task, each containing 1000 sentences, with increasing constraint complexity: $F = \{10, 100, 500, 1000\}$. We generate balanced datasets, with approximately equal support for both classes. Notice that the 1000 sentences are the same across all scenarios.

% \begin{table}[]
% \centering
% \resizebox{\columnwidth}{!}{%
% \begin{tabular}{ccc}
% \hline
% \textbf{|Forbidden Words|} & \textbf{|Valid Sentences|} & \textbf{|Invalid Sentences|} \\ \hline
% 10   & 485 & 515 \\
% 100  & 509 & 491 \\
% 500  & 518 & 482 \\
% 1000 & 477 & 523 \\ \hline
% \end{tabular}%
% }
% \caption{Support of valid and invalid sentences for the four instances of \task}
% \vspace{-0.5cm}
% \label{tab:dataset}
% \end{table}

%This problem differs from standard \textbf{keyword spotting} or \textbf{exact match filtering}, as it requires substring matching while avoiding false alerts due to synonyms or unrelated words. It has applications in content moderation, policy enforcement, and compliance monitoring, where detecting the presence of restricted terms is crucial.

\subsection{\tool for \taskAcr}
\xhdr{Model Description} In the practical scenario of \task, we train \tool to recognize whether a sentence $S$ contains a set of words $W = \{w_1, w_2, \dots, w_n$\}. The training pipeline, summarised in Fig.~\ref{fig:model_schema}, is divided into three phases:

\textit{Phase 1}: We use a frozen pre-trained sentence encoder to obtain the initial embeddings for the sentence ($e_S$) and the words ($\{e_{w_1}, e_{w_2}, \dots, e_{w_n}\}$).

\textit{Phase 2} Next, we refine these embeddings through two learnable projection layers. The sentence embeddings are refined with a linear layer $f_\chi: e_S \to \hat{e}_S$, where $\hat{e}_S$ is the refined sentence embedding. We aggregate the word embeddings into a single refined embedding $e_{\hat{w}}$ using an attention mechanism~\cite{bahdanau2014neural}. Specifically, given the embeddings $e_{w_1}, e_{w_2}, \dots, e_{w_N}$, we compute $e_{\hat{w}}$ as: 

\[
e_{\hat{w}} = \sum_{i=1}^{N} f_\gamma(e_{w_i}) \cdot f_\lambda(e_{w_i})
\]

Intuitively, we use this aggregation layer and focus on more words simultaneously to give the model a broader understanding of the context in which the words are used. For example, with \{$W_1=$ ``mount'', ``ski''\} and $W_2=$\{``mount'', ``lake''\}, the model understands that ``mount'' belongs to both winter- and spring-like scenarios.

We train the layers $\chi$, $\gamma$, and $\lambda$ using the \textit{InfoNCE} loss~\cite{oord2018representation}, which encourages higher cosine similarities for sentences and words that appear in the same set $W$. Specifically, two sentences $S_1$ and $S_2$ from the same batch are considered positive examples if they share the same set of words, and negative otherwise.

\textit{Phase 3}: After training the encoder and projection layers, we concatenate the refined sentence embedding $\hat{e}_S$ and the word embedding $e_{\hat{w}}$ into a final embedding $e_f = \hat{e}_S \parallel e_{\hat{w}}$. This concatenated embedding is then fed into a Random Forest classifier, which determines whether the words encoded in $e_{\hat{w}}$ appear in the sentence $S$ or not.

The last two phases of the training pipeline draw inspiration from the \textit{Supervised Contrastive Loss} paper~\cite{khosla2020supervised}, and are designed to learn high-quality embeddings. 

%This representation learning effort is functional to foster downstream tasks such as clustering. For instance, with meaningful embeddings one could summarise clusters of word embeddings into higher level concepts, \eg [$w_1$=``brightness'', $w_2$=``highlight'', $w_3$=``challenge'', $w_4$=``dispute''] into [$cluster_1$=(``brightness'', ``highlight''), $cluster_1$=(``challenge'', ``dispute'')]  and then into meaningful [$e_{\hat{cluster_1}}$, $e_{\hat{cluster_2}}$].

\xhdr{Training Dataset} To train \tool, we use the remaining \textit{train} and \textit{validation} partitions from CommonGen. Since more than 80\% of the sentences contain a list of three specific words, we apply synthetic augmentation to the dataset. Given a sentence (\eg "The athlete skied a snowy mountain") with three contained words (\eg {``athlete'', ``ski'', ``mountain''}), we randomly select subsets of one (\eg {``mountain''}) or two words (\eg {``athlete'', ``ski''}). The original sentence remains a valid positive sample for each subset. This enhancement allows the model to learn from training examples with varying numbers of words contained, enhancing its generalizability. As we further discuss in Sect.\ref{sec:disc_limit}, note that such augmentations, which exploit logical dependencies, are not specific to this task but generalise across various fields. For example, returning to the example in Fig.\ref{fig:figure_1}, adopting the appropriate behaviour (\eg ``inviting a Muslim for tea rather than beer'') not only aligns with the task ``How to be a good visitor'' but is also consistent with ``How to effectively socialize'' and ``How to spend quality time with locals while travelling''.

Eventually, the final dataset contains $\sim220$k labelled examples of sentences and contained words.
%%%%%%%%%%%%%%%%%%%%%%%%%%%%%%%%%%%%%
\section{Experiments}\label{sec:results}
In this section, we present the results of traditional \tss and \tool in \task. While we provide some qualitative insights, our primary focus is on reporting \textit{quantitative metrics} (e.g., accuracy, precision, and recall). A more detailed qualitative analysis, including an examination of specific model responses, can be found in Appendix~\ref{sec:appendix}.
\subsection{Experiments Settings}
%TABLE FOR NEXT SECTION: PUT HERE SO THAT WE RESPECT |PAGES|
\begin{table*}[!t]
\centering
%\vspace{-6mm}
\caption{Results of DeepSeek-R1-Distill-Llama-8B using different Test Steering Strategies as the number of forbidden words $|F|$ increases. The proposed \tool significantly outperforms other \tssAcr methods.}
 \vspace{2mm}
\resizebox{\linewidth}{!}{%
\begin{tabular}{cccccccclccc}
\hline
\multirow{2}{*}{\textbf{\tss}} & \multicolumn{3}{c}{\textbf{$|F|$: 100}} & \multicolumn{1}{l}{} & \multicolumn{3}{c}{\textbf{$|F|$: 500}} &  & \multicolumn{3}{c}{\textbf{$|F|$: 1000}} \\ \cline{2-4} \cline{6-8} \cline{10-12} 
 & \textbf{Acc.} & \textbf{Rec.} & \textbf{Prec.} & \textbf{} & \textbf{Acc.} & \textbf{Rec.} & \textbf{Prec.} &  & \textbf{Acc.} & \textbf{Rec.} & \textbf{Prec.} \\ \hline
\textbf{Simple Prompt} & 86.99 & 97.25 & 81.01 &  & 70.51 & \textbf{87.62} & 66.33 &  & 62.14 & 82.98 & 57.52 \\
\textbf{\CoT} & \textbf{87.70} & \textbf{94.16} & 83.88 &  & 68.20 & 87.12 & 63.03 &  & 59.90 & 78.34 & 56.83 \\
\textbf{Best of 3} & 85.60 & 94.16 & 80.94 &  & 62.70 & 83.30 & 58.81 &  & 58.40 & 80.16 & 55.46 \\ \hline
\textbf{\tool} & 87.50 & 79.18 & \textbf{95.76} & \textbf{} & \textbf{79.30} & 81.69 & \textbf{77.78} &  & \textbf{72.80} & \textbf{84.01} & \textbf{68.26} \\ \hline
\end{tabular}%
}
% \vspace{-1mm}
\label{tab:tss_vs_model}
\end{table*}

\xhdr{LLMs Inference} To deploy the LLMs in our \task experiments, we use \textit{SGLang}\footnote{\url{https://docs.sglang.ai/index.html}}, an open-source framework that facilitates efficient model downloading and deployment. Specifically, we select four models from SGLang’s library: \textit{Meta-Llama-3.3-8B-Instruct} and \textit{Meta-Llama-3.3-70B-Instruct} from the LLaMA family~\cite{grattafiori2024llama3herdmodels}, as well as the more recent \textit{DeepSeek-R1-Distill-Llama-8B} and \textit{DeepSeek-R1-Distill-Llama-70B} from DeepSeek~\cite{deepseekai2025deepseekr1incentivizingreasoningcapability}.
The deployment of the 70B models required four NVIDIA RTX A6000 GPUs, whereas the 8B models ran efficiently on a single A6000 GPU. When prompting the models, we set the \textit{temperature} $t$ to 0.2 for the Simple Prompt strategy and increase it to 0.4 for more sophisticated \tssAcr. The exact prompts used are provided in Appendix~\ref{sec:appendix}.

When using the Best of N strategy, we set N=3.

\xhdr{Training \tool} 
For the contrastive loss training of \tool, we perform a hyperparameter search using 4-fold cross-validation ($K=4$), ensuring that all examples sharing the same word list are assigned to the same fold to prevent data leakage. We explore embedding sizes $\{64,128,256,512\}$, learning rates $\{1e^{-4},2.5e^{-4},5e^{-4}\}$, and InfoNCE loss temperatures $\{0.05,0.1,0.2\}$, training for 30 epochs. The best configuration, determined by averaging validation results, consists of an embedding size of 128, a learning rate of $2.5e^{-4}$, a temperature of 0.05, and 24 training epochs, using \textit{all-mpnet-base-v2}\footnote{\url{https://huggingface.co/sentence-transformers/all-mpnet-base-v2}} as the pre-trained encoder. After selecting the best encoder, we train a random forest where each sentence is paired with a positive (words contained in the sentence) and a negative example (words not contained). A hyperparameter search yields an optimal configuration of 200 trees, a maximum depth of 10, and a minimum of 3 samples per leaf.

\xhdr{Metrics} Since \task is a standard binary classification problem, we evaluate performance using \textit{accuracy} (overall correctness), \textit{precision} (the proportion of predicted positive sentences that actually contain at least one forbidden word), and \textit{recall} (the proportion of actual positive sentences correctly identified). Additionally, for invalid sentences, we assess the model’s parsing ability. To do so, we introduce \textit{parsing precision} and \textit{parsing recall}. For example, given the sentence “The athlete skied the snowy mountain,” the set of forbidden words \{snow, mountain, ski\}, and the model’s prediction \{snow, ski, sun, fun\}, the parsing recall is 0.66 (2 out of 3 correct words retrieved), while the parsing precision is 0.5 (2 out of 4 predicted words are correct).

\subsection{Results}
\begin{figure}[!t]
    \centering
    % \vspace{-3mm}
    \includegraphics[width=.5\linewidth]{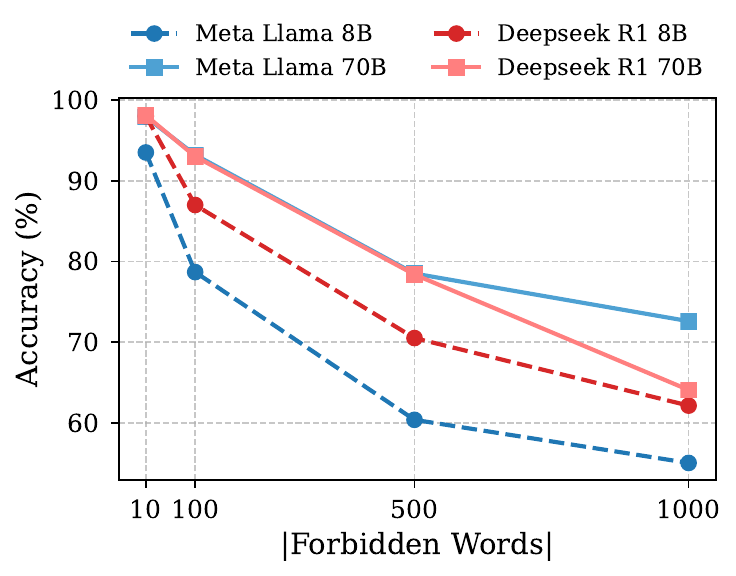}
    \vspace{-3mm}
    \caption{Accuracies with a ``Simple Prompt'' strategy as the number of forbidden words increases.}
    \vspace{-0.1cm}
    \label{fig:accuracies_vs_models}
\end{figure}

\xhdr{Is \task challenging?} We assess the effectiveness of a simple prompting strategy and find that all models, regardless of family or size, experience a roughly 30\% accuracy drop as the number of forbidden words increases from 10 to 1000 (see Fig.~\ref{fig:accuracies_vs_models}). In addition (full table on Appendix), more forbidden words lead to an increase in false alarms. For example, with 100 forbidden words, LLama 70B has a recall of 97\% and precision of 99\%, but with 1000 forbidden words, the recall only decreases to 92\%, while the precision drops to 65\%.
These results show that, despite simplicity, \task remains challenging for basic prompting strategies, suggesting that more advanced \tss are needed.

%\subsection{\tool vs traditional \tssAcr}
% PUT HERE, SO THAT WE RESPECT THE NUMBER OF PAGES
\begin{figure*}[!t]
    \centering
    %\vspace{-10mm}
    \includegraphics[width=\linewidth]{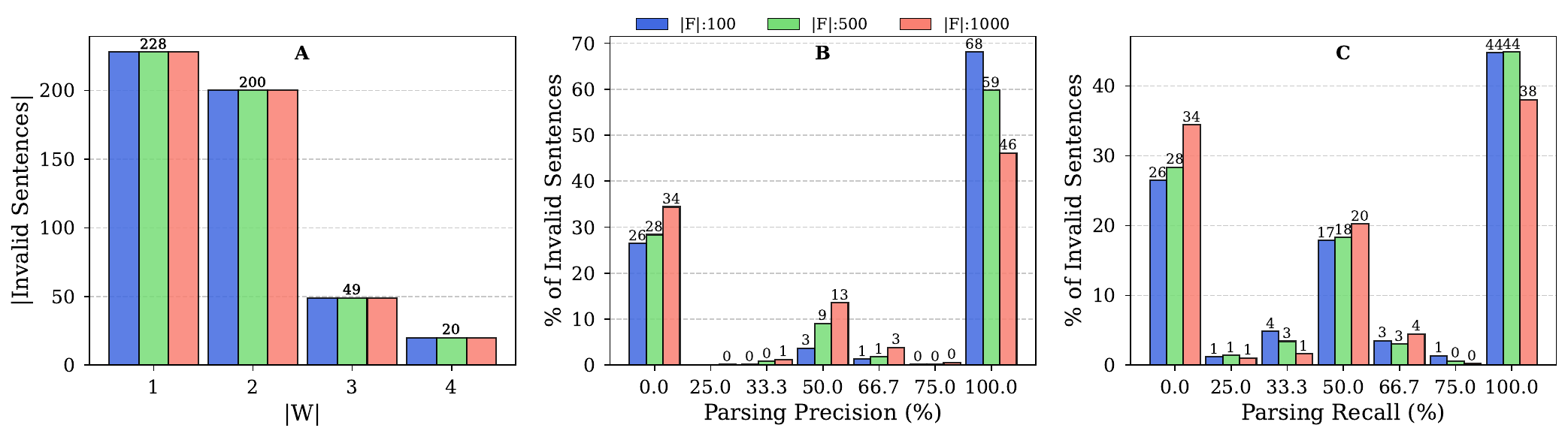}
    \vspace{-6mm}
    \caption{Analysis of recalls and precisions of \tool per invalid sentences }
    \vspace{-10mm}
    \label{fig:both}
\end{figure*}

\xhdr{\tool vs. Traditional \tssAcr Limitations} We assess the impact of advanced \tss, like \CoT and Best of 3, on \task using Deepseek's R1-8B model and compare the results with \tool.

Observe the results of Tab.~\ref{tab:tss_vs_model}. With 100 forbidden words, all methods show similar accuracy. Traditional \tssAcr has better recall, while \tool is more precise. \CoT provides minimal improvement over Simple Prompt, suggesting that the LLM is already following a "Think Step by Step" strategy. The Best of 3 strategy does not help, as, for this simple task, too many opinions lead the final LLM to overthink -- even more accentuated in the following scenario. Despite this, the LLM performs adequately in this case, which serves as our reference as we further increase the number of forbidden words.

With 500 forbidden words, the recall is similar for both traditional \tss and \tool, but \tool achieves +9\% higher accuracy due to its better precision. Both \CoT and Best of 3 degrade the performance of Simple Prompt. We find that forcing the model to reason more in simple tasks hinders its performance, as the LLM enters repetitive loops, leading to issues such as: i) confusion between its thought process and the original task, ii) overthinking (\eg, ``Should I accept synonyms?'' or ``Do plurals count?''), and iii) hallucination of non-existent words.
Contrarily, by focusing on smaller subsets of relevant words (3 for 100 forbidden words, 14 for 500, 30 for 1000), \tool helps the LLM stay on task and reduce false alarms while maintaining a good recall. 

Eventually, with 1000 forbidden words the issues observed in the 500-word case are amplified, and traditional \tss only performs 10\% better than random guessing -- remember that the problem is balanced. Although \tool performance also declines, it still performs similarly to the 70B-Llama model (~68\% precision for \tool vs ~66\% for Llama), which is promising given the $\sim10$ times smaller LLM we used here.

\xhdr{Parsing skills of LLM + \tool} Lastly, we conduct a deeper evaluation of our solution, utilizing \tool to enhance the LLM's performance. While the original task was a binary classification -- determining whether a sentence was valid or invalid -- we now refine our analysis with a more granular approach. Specifically, for invalid sentences, we assess parsing precision by measuring the proportion of predicted words that are actually present in the sentence. Additionally, we evaluate parsing recall by examining how many of the true forbidden words ($W$) the LLM correctly identifies.

Our analysis focuses on approximately 500 \textit{invalid sentences}, meaning sentences that contain at least one forbidden word ($|W|\geq1$). This selection allows us to evaluate the detector’s ability to identify relevant anomalies.

The results are shown in Fig.~\ref{fig:both}, with subfigures B and C providing key insights. These subfigures plot the percentage of invalid sentences (\textit{y-axis}) against parsing precision and recall (\textit{x-axis}). For example, they show that when using the list of relevant words identified by \tool, the LLM achieves a parsing precision of 100\% for 68\% of invalid sentences. Both distributions exhibit a trimodal pattern, with peaks at 0\%, 50\%, and 100\%. This pattern arises because most invalid sentences in the test dataset contain either one or two forbidden words (as seen in subfigure A).

Although the number of ``perfect predictions'' (both precise and accurate) consistently exceeds the number of ``bogus predictions'' (0\% precision and recall), increasing the number of candidate words ($|F|$) negatively impacts performance. Notably, the scenarios with $|F|=100$ and $|F|=500$ contain the same set of invalid sentences. This means that the true forbidden words ($W$) in these sentences remain unchanged. For them, \tool always makes the same predictions, irrespective of $F$. However, as the pool of forbidden candidate words ($F$) grows, \tool may introduce false positives into the list of relevant words returned to the LLM. These false alarms mislead the LLM, causing it to make more mistakes, thereby reducing overall performance.

% \begin{figure*}[htbp]
%     \centering
%     \begin{minipage}{0.49\linewidth}
%         \centering
%         \includegraphics[width=\linewidth]{images/05_results/recalls_per_sentence.pdf}
%         %\caption{Recalls}
%         \label{fig:fig1}
%     \end{minipage}
%     \hfill
%     \begin{minipage}{0.49\linewidth}
%         \centering
%         \includegraphics[width=\linewidth]{images/05_results/precisions_per_sentence.pdf}
%         %\caption{Precisions}
%         \label{fig:fig2}
%     \end{minipage}
%     \caption{Recalls and Precisions of \tool per Invalid Sentences }
%     \label{fig:both}
% \end{figure*}
%%%%%%%%%%%%%%%%%%%%%%%%%%%%%%%%%%%%%
\section{Conclusions}\label{sec:conclusion}
This paper introduces \problemLong, a new constrained generation problem where Large Language Models (LLMs) must adhere to a large number of constraints. We designed \task as a controlled testbed of \problemAcr in which the model classifies sentences as valid or invalid based on an increasingly large list of forbidden words.

Our experiments evaluated models from various families and sizes, testing traditional \tss and introducing \tool, a customizable support module for LLMs. The results highlight a significant performance drop across all models as the number of constraints increases. Standard \tssAcr approaches not only fail to mitigate this decline but often lead models to overthink and hallucinate constraints. In contrast, \tool proves to be the most resilient, consistently improving constraint adherence by narrowing the model’s focus.

Despite \tool’s own limitations, its effectiveness in reducing failure rates suggests a promising direction for addressing \problemAcr. With its simplicity and strong initial results, this study lays the groundwork for future research in constraint-aware LLM reasoning. By defining \problemAcr and offering open-source implementations of \task and \tool, we aim to inspire the community to explore and benchmark solutions to this critical challenge.
%%%%%%%%%%%%%%%%%%%%%%%%%%%%%%%%%%%%%
\section*{Limitations}\label{sec:disc_limit}
Here we outline the limitations of the present work.

First, while we provide examples of alternative use cases, we focus solely on a specific instance of \problem, namely \task. To better isolate the impact of an increasing number of constraints, we deliberately designed \task to minimize the role of the LLM reasoning. Although we believe that this problem has been largely overlooked in prior research, our analysis remains partial, addressing only the complexity of scenarios involving: i) multiple constraints and ii) constraints that require interpretation.

Second, our proposed model, \tool, relies on sufficient task-specific data to perform well. This dependency may limit the applicability of \tool in scenarios where task data are scarce. In the paper, we suggested that augmenting existing datasets through contrastive loss and logical dependencies between constraints and input could mitigate this issue. Additionally, as a task-specific model, \tool does not require extensive generalization, and minor "benign overfitting" is acceptable. Future work should further explore the trade-off between data availability and performance, possibly extending the analysis to contexts beyond \task.

Moreover, while we present \tool as a generic add-on module for LLMs, its architecture has only been evaluated within the \task context. More research is needed to assess its generalizability and explore how different weight architectures might affect its performance.

Finally, our work has concentrated solely on textual constraints. However, in many real-world tasks, constraints may span multiple modalities~\cite{chi2024llama, inan2023llama}. Future research could address the challenges posed by the large number of constraints in different modalities. In this regard, \tool could offer valuable flexibility, as it could be adapted with modality-specific architectures to better address these challenges.
%%%%%%%%%%%%%%%%%%%%%%%%%%%%%%%%%%%%%
\bibliographystyle{tmlr}
\bibliography{main}
%%%%%%%%%%%%%%%%%%%%%%%%%%%%%%%%%%%%%
\appendix
\section{Appendix}\label{sec:appendix}
\subsection{LLM prompts}

We here provide the prompts we used for the LLM inference:

\xhdr{Simple Prompt}

\begin{prompt}
Check if the following sentence contains one of the following set of words. 
Only answer True or False. Enclose your final answer into `<answer></answer>'.
   
For instance, if the sentence contains one of the words, answer:
- <@\texttt{<answer>True</answer>}@>; 
- <@\texttt{<answer>False</answer>}@> otherwise.
\end{prompt}

\xhdr{Chain of Thought}
\begin{prompt}
Check if the sentence contains one of the following set of words. 
Only answer True or False. Please, make sure to <@\textbf{think step by step}@>. 
Enclose your final answer into <@\texttt{<answer></answer>}@>.
   
For instance, if the sentence contains one of the words, answer:
- <@\texttt{<answer>True</answer>}@>; 
- <@\texttt{<answer>False</answer>}@> otherwise.
\end{prompt}

\xhdr{Best of 3} 

\begin{prompt}
<@\textbf{For the judges}@>: Check if the following sentence contains one of the following set 
of words. Do not include your reasoning process in the answer; Provide a short 
explanation (at most 100 words) to justify your answer. Conclude your sentence 
with <@\texttt{<answer>your answer</answer>}@>, where your answer is either <@\texttt{True}@> or <@\texttt{False}@>.
\end{prompt}

\begin{prompt}
<@\textbf{For the final verdict}@>: Give me your final opinion over the verdicts of a jury of 
<@\texttt{args.n\_rounds}@> LLMs. When prompted the following message: <@\texttt{<message>}@>, a jury of LLMs 
answered: <@\texttt{<answer>}@>. What is your final verdict? Enclose your final answer into
<@\texttt{<answer></answer>}@>.
   
For instance, if the sentence contains one of the words, answer:
- <@\texttt{<answer>True</answer>}@>; 
- <@\texttt{<answer>False</answer>}@> otherwise.

\end{prompt}

\subsection{Examples of LLMs answers}
Here we report some of the erroneous answers the LLM might return. 

\xhdr{Hallucinations} Here, the model thinks that the word ``dog'' is present in the list of forbidden words when processing the sentence ``dressed my dog up as a pirate''. However, the word is not present -- even if the conviction of the model would suggest so.  

\begin{prompt}
<@\textbf{Hallucination}@>: Okay, so I need to figure out which words from the given list are in
the sentence: "dressed my dog up as a pirate." The task is to report all the words 
that are contained in the sentence. If none are found, I should just list an empty 
list. 

First, I'll read the sentence carefully: "dressed my dog up as a pirate." I need to 
identify all the individual words in this sentence and see which ones match any from 
the provided list.

Let me break down the sentence word by word. The first word is "dressed." I should 
check if "dressed" is in the list of words. Looking through the list, I don't see 
"dressed" there. So that's one word not present.
Next is "my." Checking the list, "my" isn't there either. Moving on.
The third word is "dog." I'll look for "dog" in the list. Hmm, I don't see it either. 
So "dog" isn't present.
Then we have "up." Scanning the list, "up" isn't listed. So that's another word not
found.
The next word is "as." Checking the list, "as" isn't there. So that's not present.

# <@\textbf{HERE THE MODEL HALLUCINATE THE EXISTANCE OF THE WORD "PIRATE"}@>
Finally, "a pirate." So "a" and "pirate." Checking the list, "a" isn't there, but 
"pirate" is in the list. Wait, let me double-check. The list includes "pirate" as one 
of the words. So "pirate" is present.
Wait, but the sentence says "as a pirate," so "a" is part of that phrase, but "a" 
isn't in the list. So only "pirate" is in the list.

Let me go through the list again to make sure I didn't miss any. The words provided 
are quite extensive, but I only found "pirate" in the sentence.

So, the only word from the list that's in the sentence is "pirate."
</think>

<answer> True </answer>
\end{prompt}

\xhdr{Ovethinking \#1} Here, the LLM, prompted to ``Think Step by Step'', initially gets the correct answer -- \ie none of the words is present in the sentence. However, it is unsure about the answer, and in a second pass it eventually self-convince that all words are present. 

\begin{prompt}
<@\textbf{Overthinking}@>: Okay, so I've got this task where I need to check which words from a 
given list are present in a specific sentence. The sentence is: "winners podium with 
numerals on the sky". The words provided are quite a long list, so I need to make sure 
I go through each one carefully.

First, I'll read the sentence again to understand the context. It mentions "winners", 
"podium", "numerals", and "sky". I need to see which of the given words match any of 
these terms or are related in meaning.

Looking at the list of words, I'll go through them one by one. Starting with 
"headlamp" - doesn't seem relevant. "Success" - maybe, but the sentence doesn't 
explicitly mention success. "Wrestle" - no, that's unrelated. "Crease" - not here. 
"Poultry" - definitely not. "Softball" - nope. "Rodeo" - no. "Bait" - not present. 
"Therapy" - doesn't fit. "Whistle" - no. "Navigate" - not in the sentence. 
"Aim" - maybe, but not directly. "Attach" - no. "Melody" - no. "Mock" - no. "Pasture"
- no. "Pass" - possibly, but not directly. "College" - no. "Header" - no. "Walker" 
- no. "Resident" - no. "Calm" - maybe, but not in context. "Tiger" - no. "Hunter" 
- no. "Ambush" - no. "Motorboat" - no. "Position" - maybe, but not clearly. 
"Squeeze" - no. "Lease" - no. "Bubble" - no. "Wrap" - no. "Bounce" - no. "Teller"
- no. "Pattern" - no. "Headquarters" - no. "Brightness" - no. "Ballroom" - no. 
"Birth" - no. "Statue" - no. "Submarine" - no. "Kiss" - no. "Indoor" - no. "Entry"-no.
"Exhibition" - no. "Proceed" - maybe, but not directly. "Prey" - no. "Puffin" - no. 
"Friend" - no. "Swoop" - no. "Asteroid" - no. "Flag" - no. 
"Hamburger" - no. "Stool" - no. "Beg" - no. "Resource" - no. "Electricity" - no.
"Backseat" - no. "Anxiety" - no. "Shadow" - no. "Lego" - no. 
"Uniform" - no. "Earn" - no. "Shepherd" - no. "Marathon" - no. "Liquid" - no. 
"Theme" - no. "Bomber" - no. "Thanksgive" - no. "Request" - no. "Resistance" - no. 
"Climb" - no. "Cantilever" - no. "Bag" - no. "Push" - no. "Projector" - no. 
"Undergo" - no. "Accompany" - no. "Characteristic" - no. "Nursery" - no. "Salt" - no. 
"Mull" - no. "Workman" - no. "Sunset" - no. "Engrave" - no. "Slice" - no. "Crop" - no. 
"Clarinet" - no. "Kill" - no. "Blouse" - no. "Purple" - no. "Disappoint" - no. 
"Segment" - no. "Shoal" - no. "Memo" - no. "Cite" - no. "Hope" - no. "Deforestation" 
- no. "Dial" - no. "Tree" - no. "Militant" - no.

# <@\textbf{BEGINNING OF SECOND PASS - WHERE OVERTHINKING HAPPENS}@>
Wait, I think I might have missed "podium". Let me check again. Yes, "podium" is in 
the sentence. So that's one word. Also, "winners" is another. "Numerals" is there too. 
"Sky" is mentioned as well. Let me make sure none of the other words are in the 
sentence. "With" is a preposition, but it's not listed as a word to check. So the
words in the sentence that match the list are winners, podium, numerals, and sky.

Wait, but the original list includes "sky" as one of the words. So I should include 
that. So the words are winners, podium, numerals, and sky.
</think>
<answer> True </answer>
\end{prompt}

\xhdr{Overthinking \#2} This is an example from the Best of 3 strategy. Two support models agree that the sentence does not contain any words. However, the third model `advocates for the devil', saying that ``adorn'' is actually a synonym for ``dress'', and therefore the sentence shall be labelled invalid. The final judge model trust this third model -- as it can provide a convincing explanation -- and (erroneously) triggers the alert.

\begin{prompt}
<@\textbf{Overthinking \#2}@>: Give me your final opinion over the verdicts of a jury of 3 LLMs.
When prompted the following message: Task: Check if the following sentence contains 
one of the following set of words. Do not include your reasoning process in the anser; 
Provide a short explanation (at most 100 words) to justify your answer. 
Conclude your sentence with <answer>your answer</answer>, where your answer is either
True or False.
Sentence: dressed my dog up as a pirate .
Words: <@\texttt{<LIST\_OF\_WORDS>}@>

# <@\textbf{REPORTING ANSWERS FROM JUDGES}@>
A jury of LLMs answered:

<@\textbf{Judge 0}@>: <think> Okay, so I need to check if the sentence "dressed my dog up as 
a pirate" contains any of the given words. Let me go through each word one by one to 
make sure I don't miss any.
First, the sentence is about dressing a dog as a pirate. So, the main elements are
dressing, dog, pirate. Now, looking at the list of words, I see words like "adorn," 
which means to decorate or dress up. That's a possibility because the sentence is 
about dressing the dog.

Next, I'll check the other words to be thorough. "Container" doesn't fit here. 
"Length" and "hairstyle" are unrelated. "Mountain," "lesson," "shop," "boutique," 
"glider," "iron," "steal," "orangutan," "overall," "blur," "lava," "equestrian,"
"exert," "hear," "scuba," "slice," "manufacturing," "personality," "decor," 
"cemetery," "plantation," "ear," "excite," "argue," "inscription," "cook," "deity," 
"pallet," "skyline," "hose," "chromosome," "awning," "contrail," "brand," "right," 
"medina," "catwalk," "sheer," "berth," "lamb," "complain," "impress," "ignore," 
"onion,"  "thrive," "smell," "guitarist," "effect," "header," "aim," "ask," "dome," 
"panini," "headdress," "table," "adorn," "evacuate," "sparkle," "girlfriend," 
"synchrony," "export," "tag," "puff," "school," "fiddle," "queue," "beam," "medal," 
"intensify," "distance," "alm," "bullfight," "midair," "mangrove," "bless," "wall,"
"mow," "universe," "candle," "galaxy," "reel," "soar," "worship," "sneeze," "gallop,"
"clank," "chimney," "helmet," "allow," "leather," "embrace," "iceberg," "lantern," 
"stride," "vacation," "wallow."

Wait, I already found "adorn" as a possible match. Let me make sure there's no other
word that might fit. "Adorn" is the only one that relates to dressing or decorating 
something. The sentence is about dressing the dog, so "adorn" fits perfectly. The 
other words don't seem to connect at all. So, the sentence does contain one of the 
specified words.
</think>
The sentence contains the word "adorn," which means to decorate or dress something. 
Since the sentence is about dressing a dog as a pirate, "adorn" fits the context. 
<answer>True</answer>

<@\textbf{Judge 1}@>: <think> Alright, so I need to check if the sentence "dressed my dog up as a
pirate" contains any of the given words. Let me go through each word one by one. First,
I'll read the sentence carefully. It says the user dressed their dog up like a pirate. 
So, the main actions here are dressing and being a pirate.

Looking at the list of words, I see words like 'dress' or 'clothing' aren't present. 
The sentence mentions 'dog' and 'pirate', but those aren't in the list either. I'll
go through each word from the list to make sure I don't miss anything. Words like
'massage', 'container', 'length', 'hairstyle' - none of these are in the sentence. 
'Mountain', 'lesson', 'shop', 'boutique' - nope. 'Glider', 'iron', 'steal', 
'orangutan' - still nothing. 'Overall', 'blur', 'lava', 'equestrian', 'exert', 
'hear', 'scuba', 'slice', 'manufacturing', 'personality', 'decor', 'cemetery', 
'plantation', 'ear', 'excite', 'argue', 'inscription', 'cook', 'deity', 'pallet', 
'skyline', 'hose', 'chromosome', 'awning', 'contrail', 'brand', 'right', 'medina',
'catwalk', 'sheer', 'berth', 'lamb', 'complain', 'impress', 'ignore', 'onion',
'thrive', 'smell', 'guitarist', 'effect', 'header', 'aim', 'ask', 'dome', 'panini',
'headdress', 'table', 'adorn', 'evacuate', 'sparkle', 'girlfriend', 'synchrony',
'export', 'tag', 'puff', 'school', 'fiddle', 'queue', 'beam', 'medal', 'intensify',
'distance', 'alm', 'bullfight', 'midair', 'mangrove', 'bless', 'wall', 'mow',
'universe', 'candle', 'galaxy', 'reel', 'soar', 'worship', 'sneeze', 'gallop',
'clank', 'chimney', 'helmet', 'allow', 'leather', 'embrace', 'iceberg', 'lantern',
'stride', 'vacation', 'wallow'.

After going through each word, I don't see any that match the sentence. 
The sentence uses 'dressed' and 'dog', but those aren't in the list. The closest might
be 'adorn', but that's not present either. So, the sentence doesn't contain any of
the specified words.
</think>
<answer>False</answer>

<@\textbf{Judge 2}@>: <think>
Okay, so I need to check if the sentence "dressed my dog up as a pirate" contains any
of the given words. Let me list out the words first to make sure I don't miss any. 
The words are: massage, container, length, hairstyle, mountain, lesson, shop, boutique, 
glider, iron, steal, orangutan, overall, blur, lava, equestrian, exert, hear, scuba, 
slice, manufacturing, personality, decor, cemetery, plantation, ear, excite, argue,
 inscription, cook, deity, pallet, skyline, hose, chromosome, awning, contrail, brand,
 right, medina, catwalk, sheer, berth, lamb, complain, impress, ignore, onion, thrive,
 smell, guitarist, effect, header, aim, ask, dome, panini, headdress, table, adorn,
 evacuate, sparkle, girlfriend, synchrony, export, tag, puff, school, fiddle, queue,
 beam, medal, intensify, distance, alm, bullfight, midair, mangrove, bless, wall,
 mow, universe, candle, galaxy, reel, soar, worship, sneeze, gallop, clank, chimney,
 helmet, allow, leather, embrace, iceberg, lantern, stride, vacation, wallow.

Hmm, the sentence is about dressing a dog as a pirate. So, the main elements are 
"dog," "dress," "pirate." Now, looking through the list, I don't see "dog" listed, 
so that's a point against it. But wait, the words are about checking if any of them 
are present in the sentence, not necessarily the other way around.

Wait, the sentence is "dressed my dog up as a pirate." So the words used are 
"dressed," "my," "dog," "up," "as," "a," "pirate." None of those words are in the
provided list. Let me double-check the list again. The words are all nouns, 
I think. So "dressed" is a verb, "dog" is a noun, but "dog" isn't in the list. So
the sentence doesn't contain any of the specified words. Therefore, the answer 
should be False.
</think>
<answer>False</answer>

# <@\textbf{BACK TO THE FINAL VERDICT}@>

What is your final verdict? Just answer True or False: Ensure to enclude your final
answer into <answer></answer>. For instance, if the sentence contains one of the 
words, answer <answer>True</answer>; <answer>False</answer> otherwise.

<think>\nAlright, so I need to determine if the sentence "dressed my dog up as a 
pirate" contains any of the specified words. Let me go through each word in the 
list to see if any match.\n\nFirst, the sentence is about dressing a dog as a pirate. 
The key elements are "dress," "dog," and "pirate." Now, looking at the list of words 
provided: massage, container, length, hairstyle, mountain, lesson, shop, boutique, 
glider, iron, steal, orangutan, overall, blur, lava, equestrian, exert, hear, scuba,
 slice, manufacturing, personality, decor, cemetery, plantation, ear, excite, argue,
 inscription, cook, deity, pallet, skyline, hose, chromosome, awning, contrail, 
brand, right, medina, catwalk, sheer, berth, lamb, complain, impress, ignore, 
onion, thrive, smell, guitarist, effect, header, aim, ask, dome, panini, headdress, 
table, adorn, evacuate, sparkle, girlfriend, synchrony, export, tag, puff, school,
fiddle, queue, beam, medal, intensify, distance, alm, bullfight, midair, mangrove, 
bless, wall, mow, universe, candle, galaxy, reel, soar, worship, sneeze, gallop, 
clank, chimney, helmet, allow, leather, embrace, iceberg, lantern, stride, 
vacation, wallow.\n\nAfter reviewing each word, I notice that "adorn" is present 
in the list. The word "adorn" means to decorate or dress something. Since the sentence 
is about dressing a dog, "adorn" fits the context. Therefore, the sentence does 
contain one of the specified words.\n<answer>True</answer>'
\end{prompt}

\end{document}